\documentclass[journal,twoside]{IEEEtran}  

\IEEEoverridecommandlockouts                              

\usepackage{amsmath}

\usepackage{algorithm} 
\usepackage{algpseudocode} 
\usepackage{cite}
\usepackage{url} 
\usepackage[pdftex]{graphicx}
\usepackage{gensymb}
\usepackage{color}
\usepackage{soul}
\usepackage[normalem]{ulem} 
\usepackage{etoolbox}
\usepackage{epsfig} 
\usepackage{mathptmx} 
\usepackage{times} 
\usepackage{amsmath} 
\usepackage{amssymb}  
\usepackage{graphicx}
\usepackage{epstopdf}
\usepackage{color}
\usepackage{gensymb}
\usepackage[normalem]{ulem} 
\usepackage{tabularx,ragged2e,booktabs}
\usepackage{subfig}
\usepackage{soul}
\usepackage{tabularx,ragged2e,booktabs}
\usepackage{textcomp}

\usepackage{multirow}
\usepackage{setspace}
\usepackage{yhmath}
\usepackage{bm}
\usepackage{makecell}
\usepackage{verbatim}
\usepackage{listings}
\usepackage{tabularx}
\usepackage{multirow}
\usepackage{makecell}
\usepackage{pifont}
\usepackage{cuted}
\usepackage{capt-of}
\usepackage{afterpage}
\usepackage[T1]{fontenc}
\usepackage[utf8]{inputenc}  
\graphicspath{ {figure/} }
\usepackage{amsmath,amsfonts,amssymb}

\newcommand{\beq}{\begin{equation}}
\newcommand{\eeq}{\end{equation}}
\newcommand{\bear}{\begin{eqnarray}}
\newcommand{\bears}{\begin{eqnarray*}}
\newcommand{\eear}{\end{eqnarray}}
\newcommand{\eears}{\end{eqnarray*}}
\newcommand{\bdm}{\begin{displaymath}}
\newcommand{\edm}{\end{displaymath}}
\newcommand{\lba}{\left[\begin{array}}
\newcommand{\ear}{\end{array}\right]}

\usepackage{listings}
\usepackage{eso-pic}
\usepackage{xcolor}
\usepackage{soul} 

\makeatletter

\def\ps@RALheaders{%

    \def\@oddhead{%
        \scriptsize
        \ifnum\c@page=1
            IEEE ROBOTICS AND AUTOMATION LETTERS. PREPRINT VERSION. ACCEPTED SEPTEMBER, 2026
        \else
            \ifodd\c@page
                JI \textit{et al.}: ARCGym: Benchmarking Deep Reinforcement Learning in Autonomous Robotic Colonoscopy
            \else
                IEEE ROBOTICS AND AUTOMATION LETTERS. PREPRINT VERSION. ACCEPTED SEPTEMBER, 2026
            \fi
        \fi
        \hfill\thepage
    }%

    \def\@evenhead{%
        \scriptsize
        \thepage\hfill
        IEEE ROBOTICS AND AUTOMATION LETTERS. PREPRINT VERSION. ACCEPTED SEPTEMBER, 2026
    }%

    \def\@oddfoot{}%
    \def\@evenfoot{}%
}

\makeatother
\title{
ARCGym: Benchmarking Deep Reinforcement Learning in Autonomous Robotic Colonoscopy
}
\author{Guanglin Ji, Martina Finocchiaro, Kenny Erleben,
and Hang Yin
\thanks{Manuscript received: March 5, 2026; Revised: June 8, 2026; Accepted: August 20, 2026. This paper was recommended for publication by Editor Jessica Burgner-Kahrs upon evaluation of the Associate Editor and Reviewers comments. This work is funded by the European Union, grant number 101135082. Guanglin Ji, Martina Finocchiaro, Kenny Erleben, and Hang Yin are
with the Department of Computer Science, University of Copenhagen,
Copenhagen, Denmark (e-mail: guji@di.ku.dk; martina.finocchiaro@di.ku.dk;
kenny@di.ku.dk; hayi@di.ku.dk). 
}
}
\begin{document}
\pagestyle{RALheaders}
\AddToShipoutPictureFG*{%
  \AtPageUpperLeft{%
    \put(0,-\LenToUnit{4mm}){%
      \makebox[\paperwidth][c]{%
        \parbox{0.88\paperwidth}{%
          \centering
          \fontsize{6.5}{7.5}\selectfont
          \copyright~2026 IEEE. Personal use of this material is permitted.
          Permission from IEEE must be obtained for all other uses, in any current or
          future media, including reprinting/republishing this material for advertising
          or promotional purposes, creating new collective works, for resale or
          redistribution to servers or lists, or reuse of any copyrighted component
          of this work in other works.
        }%
      }%
    }%
  }%
}
\maketitle

\thispagestyle{RALheaders}

\begin{abstract}
Simulations for learning-based autonomous colonoscopic navigation focus mainly on fully actuated capsule robots, failing to capture the contact-rich navigation of long and flexible clinical colonoscopes. We present the Autonomous Robotic Colonoscopy Gym (ARCGym), an open-source reinforcement learning environment and benchmark for image-based navigation in clinically derived deformable colon anatomies. ARCGym supports multiple types of colonoscope robots, spanning capsule robots and flexible endoscopes, with this work focusing on flexible endoscopes including magnetic-driven tip actuation and clinically used proximally translational actuation. This work includes five CT-reconstructed colons representing typical clinical scenarios, a set of clinically meaningful navigation subtasks, and unified success metrics. We introduce a reward combining depth-based lumen alignment with a lumen-visibility score to improve learning under occlusions. Experiments across tasks, robots, and anatomies show that autonomous navigation remains challenging for both magnetic-driven and proximal-insertion flexible robots, with proximal-insertion actuation remaining an open problem.
\end{abstract}


%
\IEEEpeerreviewmaketitle
\section{Introduction}
\IEEEPARstart{C}{olorectal} cancer remains a leading cause of cancer morbidity and mortality, with approximately 1.9 million new cases and more than 900,000 deaths annually, worldwide~\cite{brenner2024reduction}. As a gold standard for early detection and intervention, colonoscopy relies heavily on operator skills and often causes patient discomfort due to colon wall stretching and loop formation.

Robotic colonoscopy systems with autonomous capabilities, e.g., the autonomous navigation of the endoscope within the colon, have the potential to simplify the procedure for clinicians while reducing patient discomfort. In recent years, autonomous navigation strategies have been increasingly investigated in simulation environments, leveraging different physics-based simulation engines, including SOFA (Simulation Open Framework Architecture) ~\cite{pore2022colonoscopy}, ~\cite{corsi2023constrained}, and Unity ~\cite{tan2025safe}.
Despite these efforts, reproduction and comparison across methods are falling behind for lacking open-source platforms. Existing works almost exclusively focus on capsule robots, failing to capture contact-induced deformations arising from underactuated, proximally driven flexible endoscopes, a key challenge frequently encountered in clinical practice.
\begin{figure}[t!]
\centering
    \includegraphics[width=0.9\linewidth]{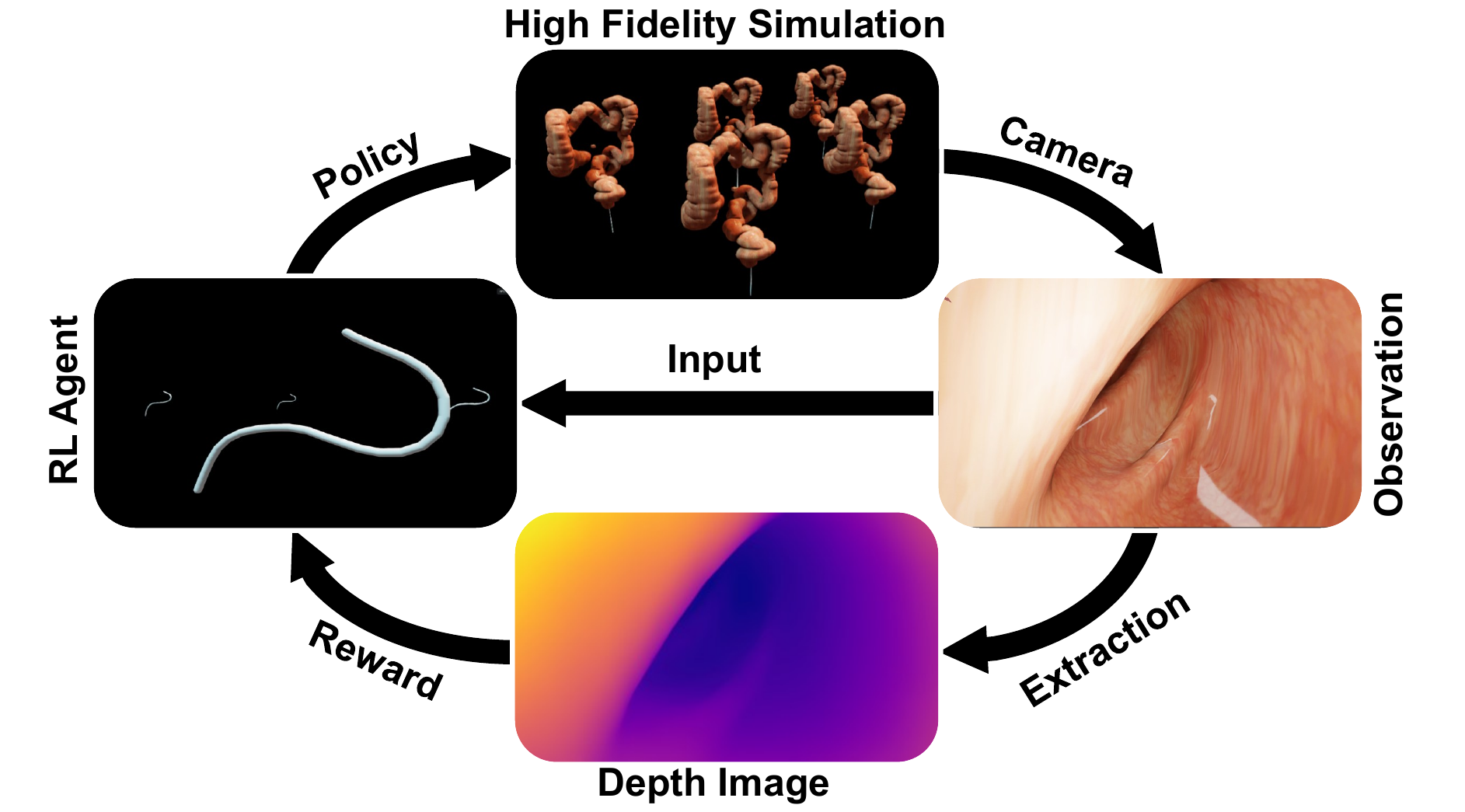}
    \captionof{figure}{Reinforcement learning framework for autonomous colonoscopy in ARCGym. An RL agent is trained to interact with the parallel deformable colon models based on the RGB observation and the reward derived from the depth image.}
    \label{fig:teaser}
\end{figure}

Clinical colonoscopy is commonly performed with a long flexible instrument whose motion is governed by distributed contact forces, friction along the colonic walls, and deformation of the colon; consequently, advancing the proximal shaft may not translate into motion at the distal tip~\cite{cheng2013modeling}. Navigation decisions depend not only on instantaneous visual observations but also on interaction history and recovery behaviors. Acute angulations, large diverticula (i.e., small pouch-like protrusions of the colonic mucosa), or suboptimal bowel preparation can cause the lumen to become temporarily invisible; in such cases, the scope motion becomes constrained by wall contacts and forward advancement should be avoided until the lumen is re-identified~\cite{cohen2022successful}. This phenomenon is much less common in capsule robots, where the robot is fully actuated and is not subject to the same contact-driven kinematic constraints.

Reinforcement learning (RL) provides a natural framework for learning closed-loop navigation directly from images, overcoming the complexity of crafting control to accommodate interaction-driven dynamics of flexible endoscopy. 
Existing medical robotic learning benchmarks have been proposed to accelerate advancements in robot-assisted laparoscopy, primarily developed for rigid manipulators and platforms based on the da Vinci Research Kit~\cite{xu2021surrol, schmidgall2024surgical, wu2024surgicai}. To the best of our knowledge, no such colonoscopy navigation benchmark has been reported.

\begin{figure*}[t!]
\centering
    \includegraphics[width=0.9\linewidth]{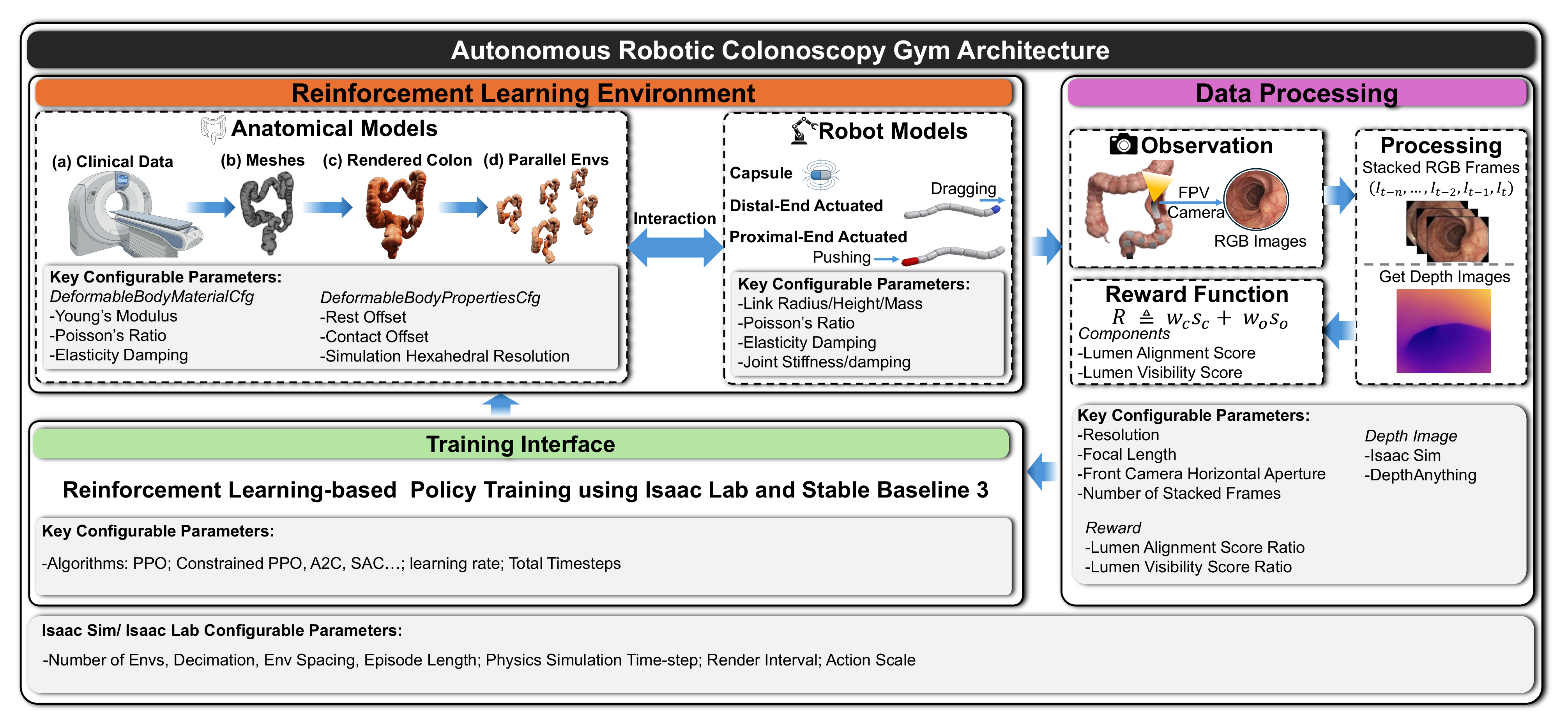}
    \captionof{figure}{Architecture of the Autonomous Robotic Colonoscopy Gym (ArcGym). (a--d) Clinical CT data are reconstructed into colon meshes, rendered soft-tissue models, and replicated into parallel simulation environments for scalable reinforcement learning training. The framework integrates deformable anatomical models and multiple robot actuation modes within a reinforcement learning environment built on Isaac Lab/Isaac Sim. Depth images are processed to compute lumen-based rewards for RL policy training.}
    \label{fig:software_architecture}
\end{figure*}
In this work, we introduce ARCGym, a benchmark environment for deep reinforcement learning in autonomous robotic colonoscopy with soft-tissue deformation, contact, and flexible endoscopes. Fig.~\ref{fig:teaser} provides an overview of the RL-based colonoscopy navigation pipeline, and Fig.~\ref{fig:software_architecture} demonstrates the software architecture of our platform. ARCGym is implemented in Isaac Lab with an RL interface, supporting reproducible comparisons under unified task definitions and metrics. Our contributions can be summarized as:
\begin{itemize}
    \item An open-source framework for RL-based navigation in robotic colonoscopy, comprising five clinically derived colon anatomies, a realistic mechanical model of deformable colonic tissue behavior, anatomically defined navigation subtasks, and standardized success metrics. The code is available at https://github.com/bentferrari/IRE\_ARCGym.
    \item Various robot types include a 6-DoF capsule, a magnetically driven distal-end actuated flexible endoscope, and a conventional proximal-insertion flexible endoscope, all sharing unified observation and action interfaces for cross-platform comparison.
    \item {\color{black}A novel reward formulation that combines six reward components to provide a more reliable learning signal across the tested navigation conditions, as reflected by the center-alignment results.}
    \item Extensive baseline and generalization evaluations across tasks and anatomies reveal navigation difficulties when using flexible endoscopes.
\end{itemize}

\section{Related Work}
\label{sec:related_work}

\subsection{Reinforcement Learning for Medical Robotics}
Reinforcement learning has been investigated to map endoscopic observations directly to control actions, reducing reliance on explicit planning and control stages.
Prior simulation-based colonoscopy navigation work mostly targets capsule robots and uses simplified interaction models~\cite{pore2022colonoscopy},~\cite{corsi2023constrained}.
Safety-oriented extensions, such as shield RL and constrained RL, were explored in~\cite{ji2021towards}, ~\cite{corsi2023constrained}, and human intervention for safer learning was studied in~\cite{tan2025safe}.
Beyond colonoscopy, multimodal learning for other flexible endoscopy settings, such as bronchoscopy has been explored~\cite{zhao2024bronchocopilot}. Similarly, RL has been explored for other gastrointestinal endoscopic behaviors beyond lumen-following as well, such as coverage scanning for wireless capsule endoscopy~\cite{zhang2022deep}. These studies primarily evaluate navigation under simplified or task-specific settings, while motion constrained by interactions between the passive shaft and deformable anatomy remains underexplored. ARCGym provides a standardized environment to study navigation under realistic interaction dynamics.

Moreover, recent medical-robot learning studies demonstrate diverse RL-driven autonomy problems. Turan~et~al.~\cite{turan2019learning} learn control policies instead of motion planning policies using an endoscopic capsule robot, which illustrates closed-loop control for the capsule robot from visual feedback. In another imaging modality, Li~et~al.~\cite{li2023rl} propose attention-augmented DRL for autonomous transesophageal echocardiography (TEE) probe guidance. Focusing on surgical skill acquisition, Tan~et~al.~\cite{tan2019laparoscopy_drl} study DRL for robot-assisted laparoscopic training. For endoscopic perception and control with richer supervision, EndoVLA~\cite{endovla2025corl} introduces a vision--language--action framework for precise autonomous tracking in endoscopy. Moving toward more general surgical autonomy, Long~et~al.~\cite{long2025surgical} present surgical embodied intelligence for generalized task autonomy in laparoscopic robot-assisted surgery. In terms of simulation, Sim4EndoR~\cite{11127627} provides an RL-centered simulation platform for endovascular robotics task automation. These works demonstrate the potential of RL in medical robotics, with a focus on specific procedures and control objectives. ARCGym instead concentrates on the navigation problem in colonoscopy.






\begin{table*}[t!]
\centering
\small
\caption{Capability Comparison Between ARCGym and Existing Platforms}
\label{tab:capability_comparison}
\renewcommand{\arraystretch}{1.25}
\setlength{\tabcolsep}{6pt}
\begin{tabular}{p{0.5\columnwidth} c c c c c c}
\toprule
\multirow{2}{*}{\textbf{Literature}} 
& \multicolumn{3}{c}{\textbf{Robot Type}} 
& \multirow{2}{*}{\textbf{Deformable}} 
& \multirow{2}{*}{\textbf{Parallel Env.}} 
& \multirow{2}{*}{\textbf{RL Benchmark}} \\
\cmidrule(lr){2-4}
& \textbf{Capsule} 
& \textbf{Long Flexible} 
& \textbf{dVRK} 
&  &  &  \\
\midrule

SurRoL~\cite{xu2021surrol}
& - & - & \ding{51}
& - & - & \ding{51} \\

Surgical Gym~\cite{schmidgall2024surgical}
& - & - & \ding{51}
& - & \ding{51} & \ding{51} \\

Sim4EndoR~\cite{11127627}
& - & \ding{51} & -
& \ding{51} & - & \ding{51} \\

LapGym~\cite{scheikl2023lapgym}
& - & - & \ding{51}
& \ding{51} & \ding{51} & \ding{51} 
\\

CathSim~\cite{jianu2024cathsim}
& - & \ding{51}& - 
& - & - & \ding{51} \\

\midrule
Deep Visuomotor~\cite{pore2022colonoscopy}
& \ding{51} & - & -
& \ding{51} & - & - \\

Constrained RL~\cite{corsi2023constrained}
& \ding{51} & - & -
& \ding{51} & - & - \\

Human Intervention RL~\cite{tan2025safe}
& - & -(Only tip) & -
& \ding{51} & - & - \\

\textbf{ARCGym (Ours)}
& \textbf{\ding{51}} & \textbf{\ding{51}} & -
& \textbf{\ding{51}} & \textbf{\ding{51}} & \textbf{\ding{51}} \\
\bottomrule
\end{tabular}
\end{table*}
\subsection{Medical Simulation Platforms and Benchmarks}
The majority of existing medical robot learning platforms are developed for laparoscopic surgical systems, while a smaller number address endovascular intervention. ARCGym instead targets flexible endoscopic navigation, where the navigation progress depends on visual perception and interaction-constrained motion in highly deformable anatomy.

SurRoL~\cite{xu2021surrol} provides an open-source RL platform compatible with the da Vinci Research Kit (dVRK). Surgical Gym~\cite{schmidgall2024surgical} proposes a high-performance GPU-based RL platform for surgical robots. SurgicAI~\cite{wu2024surgicai} introduces a hierarchical platform for fine-grained surgical policy learning and benchmarking. Similarly, LapGym~\cite{scheikl2023lapgym} provides an open-source framework for reinforcement learning in robot-assisted laparoscopic surgery. Earlier open-sourced RL environments for surgical robotics have been discussed in~\cite{richter1903open}. Complementary toolkits aim to standardize interfaces and enable safer evaluation in laparoscopic robots. AMBF-RL~\cite{varier2022ambf_rl} provides a real-time simulation-based RL toolkit for medical robotics. ORBIT-Surgical~\cite{yu2024orbit} proposes an open-simulation framework for learning surgical augmented dexterity. These platforms focus on manipulation and dexterity learning with rigid instruments. In contrast, ARCGym complements existing surgical benchmarks and enables quantitative evaluation of learning-based navigation strategies by modeling flexible endoscopic navigation and providing standardized subtasks and anatomies.

Beyond laparoscopic platforms, endovascular robotics has received dedicated simulation support. CathSim~\cite{jianu2024cathsim} provides an open-source simulator for endovascular intervention. Building on an RL-centric design, Sim4EndoR~\cite{11127627} proposes a simulation platform for endovascular robotics task automation. These platforms address catheter steering in vascular structures, where navigation is primarily position-controlled along a known lumen. ARCGym instead focuses on vision-driven navigation in colonoscopy, where the whole structure is unknown. In Table~\ref{tab:capability_comparison}, we list the capability comparison between the existing platforms and ours.


\section{Colonoscopy Simulation}\label{sec:sim}
\subsection{Modeling of Colons in Isaac Sim}

ARCGym models the human colon uses the finite-element (FEM) framework in NVIDIA Isaac Sim, as it provides GPU-accelerated deformable body simulation with stable contact handling while supporting large-scale parallel environments required for reinforcement learning. The colon anatomies simulated in ARCGym are derived from real human colons segmented from CT colonography (CTC) images obtained from the HQColon dataset \cite{finocchiaro2026clinically}. Surface triangular meshes were generated from these segmentation and imported into NVIDIA Isaac Sim as deformable objects. In collaboration with an expert abdominal radiologist (six years of experience in clinical CTC assessment), we selected five cases representing typical clinical scenarios with varying levels of procedural difficulty:

\begin{itemize}
    \item \textbf{Colon 1:} Male, 65 years old, no polyps, diverticula in the descending colon, and a slightly redundant sigmoid colon with sharp angulations.
    
    \item \textbf{Colon 2:} Male, 51 years old, no polyps, examined in the prone position, with a redundant transverse colon.
    
    \item \textbf{Colon 3:} Same patient as Colon 2, examined in the supine position, also presenting a redundant transverse colon.
    
    \item \textbf{Colon 4:} Female, 51 years old, no polyps or diverticula, but a highly redundant sigmoid and transverse colon with multiple sharp bends.
    
    \item \textbf{Colon 5:} Male, 60 years old, 16 polyps, diverticula in both the ascending and descending colon, and a redundant sigmoid colon. The patient also suffers from constipation and presents multiple polyps.
\end{itemize}

Colons 1 and 5 include diverticula, which can increase navigation complexity because they may be mistaken for the lumen. Redundancy in the sigmoid and transverse colon further increases technical difficulty due to looping and sharp angulations. The case with multiple polyps simulates a high-risk patient, potentially with colorectal cancer. Finally, the same patient was modeled in two different positions (prone and supine) to assess the transferability of navigation policies under positional changes.

{\color{black}The colon is modeled in a fully insufflated configuration as a nearly incompressible elastic body with a Young’s modulus of $5\mathrm{MPa}$~\cite{massalou2019mechanical} and a Poisson ratio of 0.49~\cite{Tian2023BowelCancer}. Human-colon tensile measurements report MPa-scale elastic moduli after the initial toe-in region, with stiffness varying with loading direction and rate~\cite{massalou2019mechanical}; we therefore treat $5\mathrm{MPa}$ as an effective stiffness for the distended colon model rather than a small-strain tissue modulus. Material sensitivity is further evaluated in Section~\ref{sec:material_sensitivity}.}

To approximate anatomical fixation and prevent global drift, a sparse kinematic constraint scheme based on manual anchor selection is employed. Vertices corresponding to the rectum, descending and ascending colon, and the two colonic flexures were manually selected and fixed to approximate the anatomical constraints of the human colon~\cite{loeve2013mechanical}. These constraints leave, among other parts, the sigmoid and transverse colon less constrained, as these segments are primarily responsible for loop formation during colonoscopy. This approach provides stable boundary conditions without compromising realistic tissue deformation.

\subsection{Modeling of Endoscopes in Isaac Sim}
ARCGym covers the most commonly studied robotic categories in colonoscopy literature: a capsule robot, a magnetic-driven flexible colonoscope with a fully actuated distal tip, and a conventional flexible colonoscope with proximal insertion. 

\subsubsection{Capsule Robot}
The capsule endoscope is modeled as a single rigid body with six-degree-of-freedom (6-DoF) motions. Its dimensions (diameter $10~\mathrm{mm}$, length $20~\mathrm{mm}$, mass $5~\mathrm{g}$) lie within the typical size range of clinical capsule endoscopes~\cite{6041014}. The capsule robot has been widely investigated in the literature~\cite{pore2022colonoscopy},~\cite{corsi2023constrained},~\cite{tan2025safe}; therefore, we provide it in the open-source code and focus on flexible-colonoscope navigation in this paper.

\subsubsection{Flexible Endoscope}
The flexible endoscopes are modeled as compliant articulation bodies discretized into rigid cylindrical links connected by alternating-axis revolute joints, a common pseudo-rigid link model for continuum robots~\cite{jones2006kinematics}. One endoscope uses 6-DoF distal-tip velocity control to represent a magnetically driven robot~\cite{martin2020enabling}; the other uses proximal insertion and axial rotation with distal pitch-yaw steering to approximate conventional clinical colonoscope operation.



\begin{figure}[t!]
    \centering
    \includegraphics[width=0.85\linewidth]{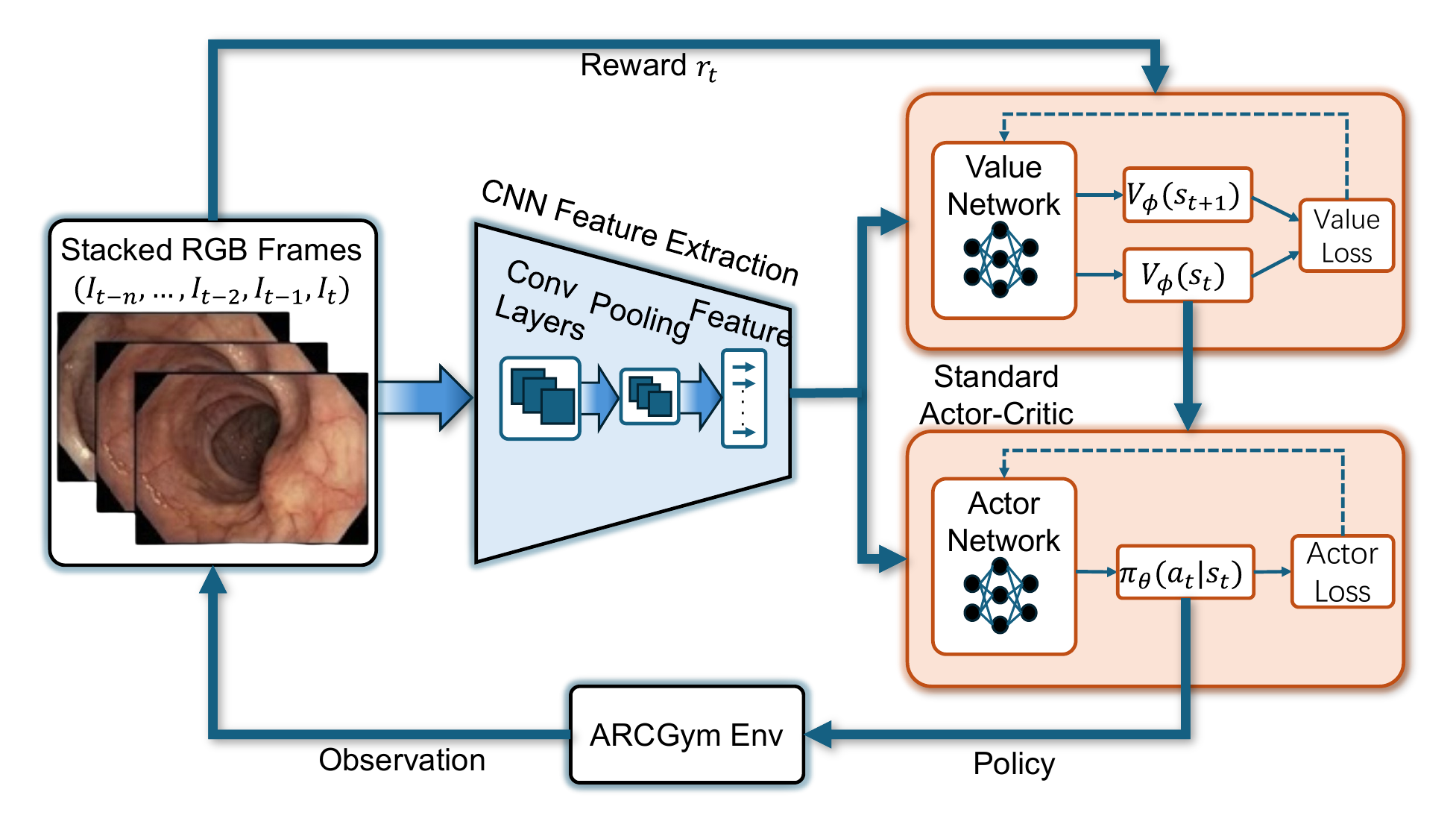}
    \caption{Vision-based actor--critic learning pipeline in ARCGym. The agent observes stacked monocular RGB frames $(I_{t-n}, \ldots, I_{t})$, which are encoded by a CNN feature extractor and fed into a standard actor--critic architecture.
}
    \label{fig:rl_framework}
\end{figure}
\subsection{ARCGym RL Library}
Fig.~\ref{fig:rl_framework} illustrates the vision-based actor-critic learning pipeline used in ARCGym.

\subsubsection{Observation Space}
The full states of either the capsule or the flexible endoscope are generally unavailable in practice due to limited sensing capability. To approximate realistic colonoscopy perception, the observation space is defined as the RGB image captured from the endoscopic camera for all robots. {\color{black}Depth images are not included in the policy observation. They are used only during simulation training to compute the reward. Therefore, the learned policy acts from RGB observations only, while the depth signal serves as a simulator-derived dense training signal.}

\subsubsection{Action Space
}
The capsule robot and the magnetic-driven flexible endoscope are both fully actuated with six degrees of freedom (DoF) motion and share the same action space
$v_1\triangleq[v_x, v_y, v_z, \omega_p, \omega_y, \omega_r]^T\in\mathbb{R}^{6}$, where $v_x,v_y,v_z$ denotes
translations along the left--right, up--down, and forward--backward axes, and
$\omega_p,\omega_y,\omega_r$ denotes pitch, yaw, and roll angular velocities,
respectively. The action space for the proximal-insertion endoscope is $v_2\triangleq[v_z, \omega_p, \omega_y, \omega_r]^T\in\mathbb{R}^{4}$.

\subsubsection{Reward Function}
{\color{black}Although the policy observes only RGB images, the simulator provides depth images during training for reward computation.} Given a depth image $D\in\mathbb{R}^{H\times W}$, we define a deep-region mask by thresholding the depth values:
\begin{equation}
\mathcal{S}_{\text{d}}\triangleq\{(r,c)\mid D(r,c)\ge \tau\},
\label{eq:deep_region_set}
\end{equation}
where $\tau=\beta\max(D)$, and $\beta\in(0,1]$ control the depth threshold. The centroid of the region of interest (ROI) is computed as
\begin{equation}
\mathbf{c}\leftarrow(\bar r,\bar c)\leftarrow\left(
\frac{1}{|\mathcal{S}_{\text{d}}|}\sum_{(r,c)\in\mathcal{S}_{\text{d}}} r,\quad
\frac{1}{|\mathcal{S}_{\text{d}}|}\sum_{(r,c)\in\mathcal{S}_{\text{d}}} c
\right).
\label{eq:deep_centroid}
\end{equation}

Let the image center be $\mathbf{c}_0$, then the deepest-region deviation and its maximum possible value are
\begin{equation}
r\triangleq\|\mathbf{c}-\mathbf{c}_0\|_2,
\qquad
r_{\max}\triangleq\left\|\left[\frac{W-1}{2},\,\frac{H-1}{2}\right]^\top\right\|_2.
\end{equation}
We define the normalized centering term as
\begin{equation}
s_c \triangleq 1-\frac{r}{r_{\max}}.
\end{equation}

We define a lumen visibility score to explicitly penalize these scenarios and encourage lumen-facing behaviors:
\begin{equation}
s_1\triangleq\mathrm{clip}\left(\frac{d_{\max}-\tau_w}{\tau_\ell-\tau_w},0,1\right),\
s_2\triangleq\mathrm{clip}\left(\frac{\rho_d-\rho_{\min}}{1-\rho_{\min}},0,1\right),
\end{equation}
\begin{equation}
s_3\triangleq\mathrm{clip}\left(\frac{\Delta d-\delta_{\min}}{\delta_{\mathrm{good}}-\delta_{\min}},0,1\right),\nonumber
\end{equation}
where $d_{\max}\triangleq\max(D)$ and define the depth range as
$\Delta d \triangleq d_{\max}-d_{\min}$.
The operator $\mathrm{clip}(x,0,1)\triangleq\min(\max(x,0),1)$ rescales a scalar into $[0,1]$.
The normalized lumen visibility score is defined as
\begin{equation}
s_o \triangleq 2( \beta s_1 + (1-\beta)s_2 )\, s_3-1.
\end{equation}
%
%
Then, the total reward is defined as a weighted sum:
\begin{equation}
R \triangleq w_c s_c + w_o s_o, 
\end{equation}
where the total reward $R$ for each timestep is bounded by $(-1, 1)$. All depth images can be either obtained from Isaac Sim or Depth Anything~\cite{yang2024depth}.

\begin{figure}[t!]
    \centering
    \includegraphics[width=0.75\columnwidth]{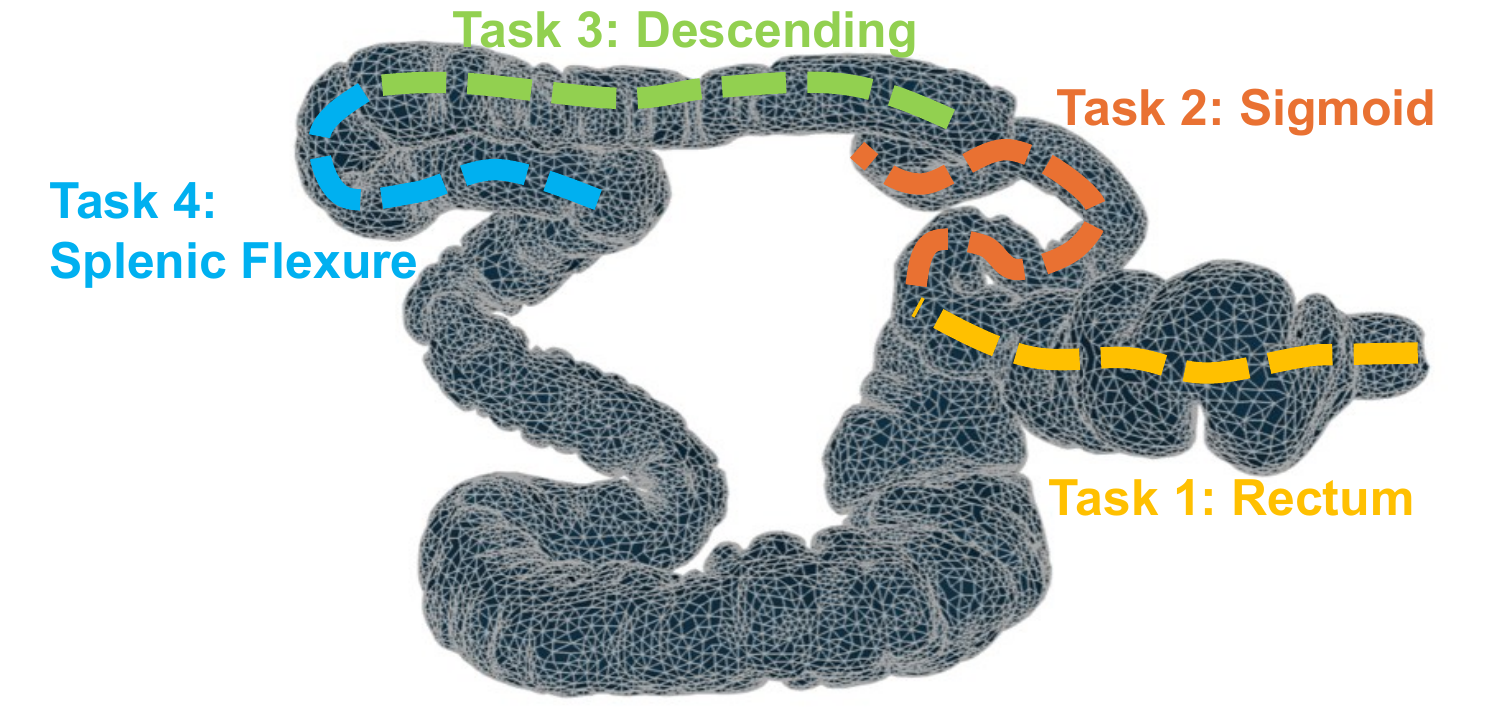}
    \caption{Definition of navigation subtasks. Task 1: Rectum navigation; Task 2: Sigmoid navigation; Task 3: Relatively Straight Part Navigation (Descending); Task 4: Turning Behavior at the Splenic Flexure. Task 2 and Task 4 with sharp turns demonstrate the most challenging navigation behavior.}
    \label{fig:task_definition}
\end{figure}
\subsection{Navigation Task Definitions}
\label{subsec:exp_tasks}

Clinical studies identify the rectosigmoid junction and splenic flexure as the main colonoscope-insertion bottlenecks: acute sigmoid and flexure angulations cause looping that prevents effective forward propagation~\cite{waye2013difficult}, and early splenic-flexure reach predicts cecal-intubation competency~\cite{mccarthy2016early}. Thus, the rectum--sigmoid--splenic segment is the principal control bottleneck, whereas later navigation is less dominated by left-colon insertion limits.

{\color{black}
Based on this evidence, ARCGym uses reaching and passing the splenic flexure as a clinically meaningful intermediate milestone. This milestone provides a representative benchmark target for evaluating whether a policy can overcome the main left-colon insertion bottlenecks. Complete cecal intubation remains an important future extension of the benchmark.
} Fig.~\ref{fig:task_definition} illustrates the task definitions.

{\color{black} The task boundaries are derived from a continuous teleoperated insertion trajectory rather than independently designed isolated stages. Starting from the rectum, the endoscope is teleoperated continuously through the colon. At the end of each anatomical segment, the current endoscope configuration is recorded as the target configuration of the current task and reused as the initial configuration of the next task. The teleoperation then continues into the following segment. Therefore, consecutive tasks preserve the sequential structure of a continuous colonoscopy procedure while providing standardized start and target states for reproducible evaluation.}

\subsubsection{\textbf{Task~1 -- Rectum Navigation}}
The robot is initialized near the rectal entry and navigates toward the sigmoid colon. Owing to the wide and approximately straight lumen, this task evaluates whether a policy can establish stable, visually aligned forward motion from a valid entry configuration.

\subsubsection{\textbf{Task~2 -- Sigmoid Navigation}}
The robot navigates the sigmoid colon from its proximal entrance to its distal end. {\color{black}The navigation in the flexible sigmoid part can produce alpha-loop-like configurations, where proximal insertion no longer maps one-to-one to distal tip advancement~\cite{waye2013difficult}. Such loop-induced motion ambiguity is not captured by rigid-tube simulators with fixed lumen geometry, but is central to operator difficulty and learning-based policy generation.}

\subsubsection{\textbf{Task~3 -- Straight-Part Navigation (Descending)}}
This task covers navigation through the descending colon, representative of relatively straight colon segments. The primary challenge arises from sustained long-horizon progression rather than sharp curvature or extreme deformation.

\subsubsection{\textbf{Task~4 -- Turning at the Splenic Flexure}}
This task focuses on navigation at the junction between the descending and transverse colon, where the lumen makes a sharp change in direction. Successful navigation requires the endoscope to adjust its orientation in time while continuing forward motion. The flexure is fixed in either clinical or simulation scenario. Therefore, compared to the sigmoid colon, the main difficulty lies in maintaining lumen alignment during turning rather than handling strong deformation.

\begin{figure}[t!]
    \centering
    \includegraphics[width=0.92\columnwidth]{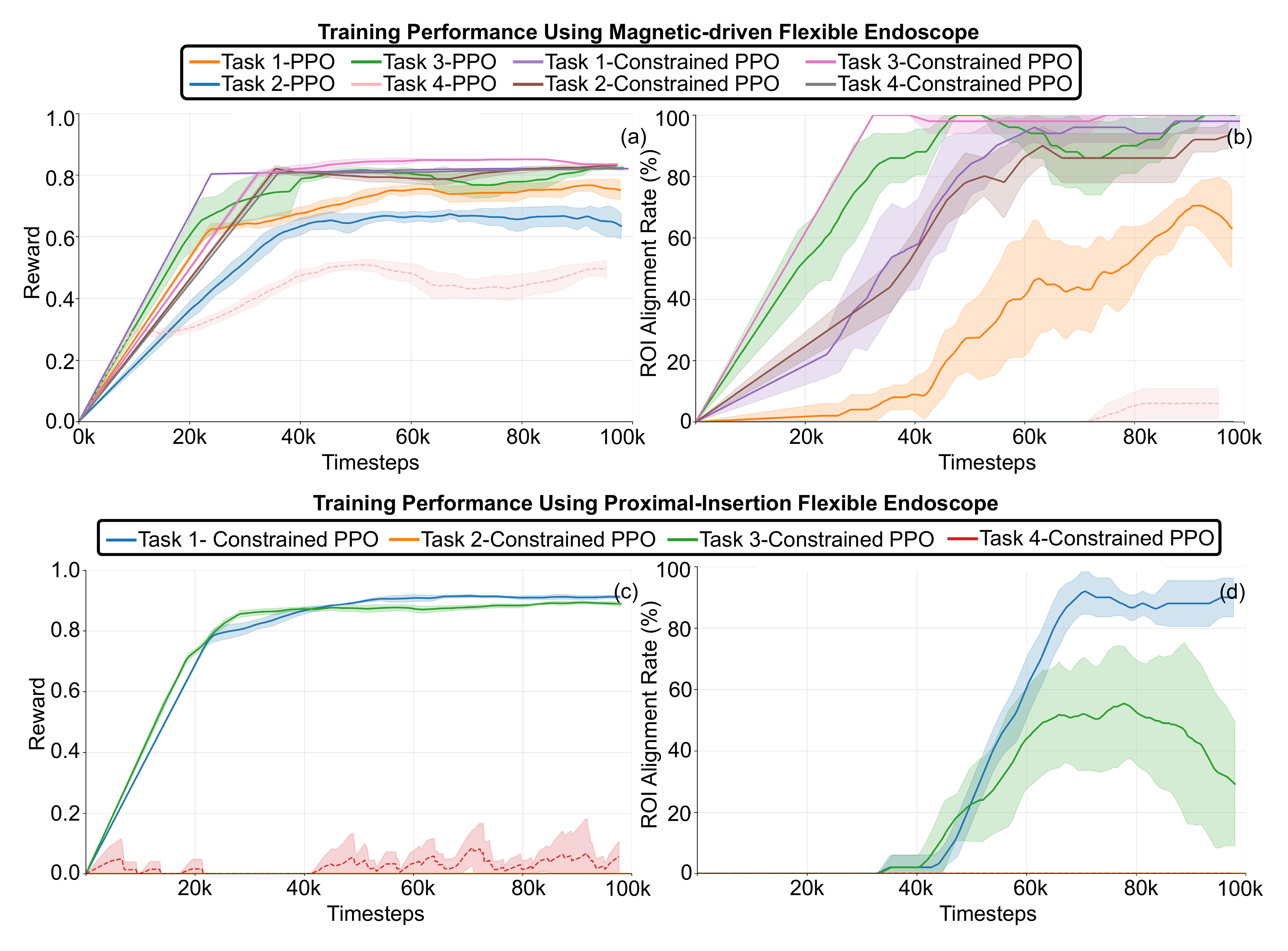}
    \caption{Normalized reward and ROI alignment rate over training timesteps for four colonoscopy navigation tasks. (a)--(b) Magnetic-driven flexible endoscope. (c)--(d) Proximal-insertion flexible endoscope. Results indicate that Task 2 and Task 4 are the most challenging overall. Under the proximal-insertion endoscope, (c) and (d) demonstrate a learnable behavior, although the goal reaching success rate is reported all zeros in Table~\ref{tab:rl_results}.}
    \label{fig:training_curves}
\end{figure}

\begin{table}[!t]
\begingroup
\color{black}
\caption{Training throughput. Steps denote
aggregate environment transitions across all parallel environments.}
\label{tab:parallel_throughput}

\makebox[\columnwidth][c]{%
\begin{tabular}{ccccc}
\hline
Envs & Steps & Time (min) & Steps/s & Speedup \\
\hline
1 & 44,000 & 60.00 & 12.22 & 1.00$\times$ \\
5 & 44,000 & 21.16 & 34.65 & 2.84$\times$ \\
\hline
\end{tabular}%
}

\endgroup
\end{table}

\section{Experimental Results and Discussion}
\label{sec:experiments}


\begin{table*}[t!]
\small
\centering
\caption{{\color{black}Evaluation of reinforcement learning algorithms on the designed tasks using the magnetic-driven flexible endoscope and proximal-insertion flexible endoscope. On-policy algorithms perform better, especially for constrained PPO, all tasks besides Task 2 can achieve high success rate. In contrast, the off-policy algorithm SAC fails to finish all tasks.}}
\label{tab:rl_results}
\begingroup
\color{black}
\begin{tabular}{c|cccc|c}
\toprule
Algorithm 
& \multicolumn{4}{c|}{Success Rate (\%): Distal Magnetic-driven Endoscope} 
& Success Rate (\%): Proximal-insertion Endoscope\\
\cmidrule(lr){2-6}
& Task 1 & Task 2 & Task 3 & Task 4 & All Tasks \\
\midrule

PPO 
& 20.0 & 0.0 & 40.0 & 0.0 
& 0.0 \\

Constrained PPO~\cite{corsi2023constrained}
& \textbf{100.0} & \textbf{30.0} & \textbf{90.0} & \textbf{80.0} 
& 0.0 \\

A2C
& 20.0 & 0.0 & 20.0 & 0.0 
& 0.0 \\

SAC
& 0.0 & 0.0 & 0.0 & 0.0 
& 0.0 \\

\bottomrule
\end{tabular}
\endgroup
\end{table*}
\subsection{Experimental Setup}
\label{subsec:exp_setup}
All experiments use five anatomically distinct deformable colon models and two flexible colonoscopes, since capsule robots are covered in~\cite{pore2022colonoscopy},~\cite{corsi2023constrained},~\cite{tan2025safe}. As in state-of-the-art (SOTA) RL methods for colonoscopy navigation~\cite{pore2022colonoscopy},~\cite{corsi2023constrained},~\cite{tan2025safe}, the robots move with a constant forward velocity. Policies are trained on a workstation with an Intel Core i9-10850K CPU and an NVIDIA RTX PRO 6000 Blackwell Max-Q GPU under Ubuntu 24.04.3. {\color{black}Each method and task are trained or evaluated using three random seeds. For each seed, two evaluation runs are conducted with five parallel instances of the same colon, resulting in 30 trials per condition.} {\color{black}At a matched budget of 44,000 transitions, five parallel environments achieve 34.65 steps/s versus 12.22 steps/s with one environment, corresponding to a \(2.84\times\) hardware-specific throughput improvement as shown in Table~\ref{tab:parallel_throughput}.}

\subsection{Navigation Performance and Ablation Studies}
\label{subsec:exp1}

\begin{figure}[t!]
    \centering
    \includegraphics[width=0.89\columnwidth]{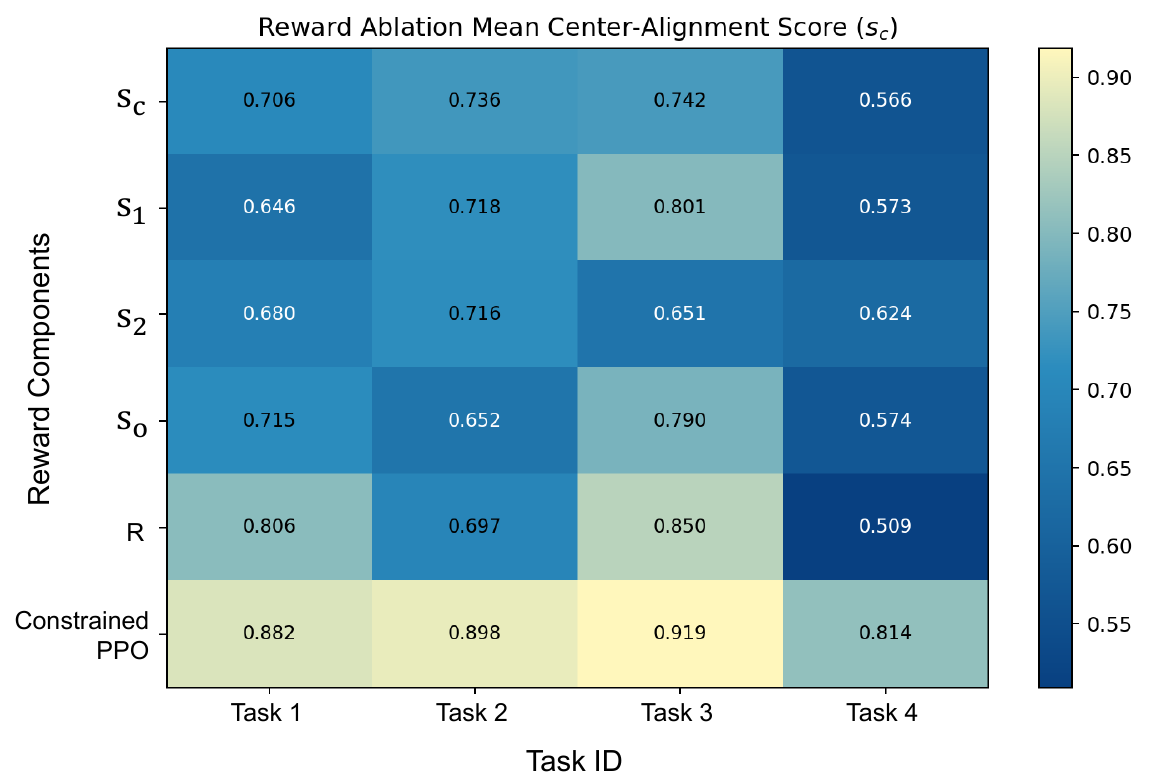}
    \caption{{\color{black}Average reward ablation results on Colon~1 and Colon~2 across Task~1--Task~4. Mean center-alignment score $s_c$ is reported for PPO with different reward terms and constrained PPO with full reward.}}
    \label{fig:test}
\end{figure}
\subsubsection{Navigation Performance Across Tasks}

We trained a distal magnetic-driven flexible endoscope to navigate through Colon 1 using four state of-the-art (SOTA) RL algorithms in colonoscopy. We report the PPO and constrained PPO in Fig.~\ref{fig:training_curves} (a)-(b), showing that Task~4 achieves the lowest reward and the slowest improvement, indicating high difficulty in aligning the ROI center, while Task~3 is the easiest due to its long and relatively straight lumen. 

By contrast, the proximal-insertion flexible endoscope learns only limited navigation behaviors, as shown in Fig.~\ref{fig:training_curves}(c)-(d). Its normalized reward and lumen alignment rate remain lower than those of the magnetic-driven endoscope for most tasks. Task~1 improves moderately and reaches a non-zero lumen alignment rate, suggesting that simple lumen geometries can still be handled under proximal-insertion actuation. However, Task~2, Task~3, and Task~4 do not converge, with alignment rates remaining close to zero. This gap reflects the harder contact-driven control problem under proximal insertion, especially in sharp bends or long-range shape propagation where proximal pushing does not directly move the distal tip. 

\begin{figure}[t!]
    \centering
    \includegraphics[width=0.75\columnwidth]{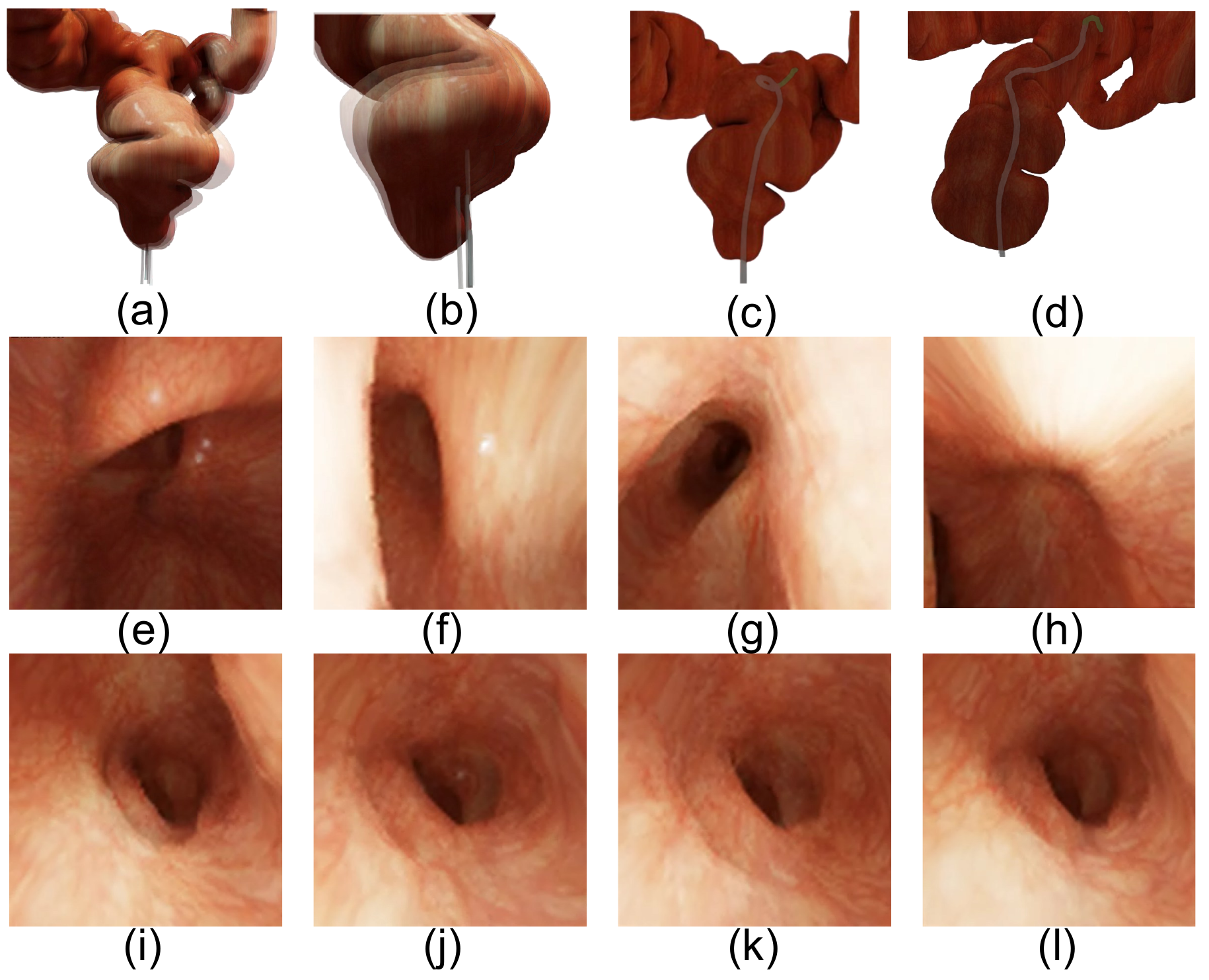}
    \caption{Simulation results. (a)-(b) Deformation caused by interaction; (c)-(d) Proximal-insertion and $\alpha$ loop formation; (e) Rectum; (f) Sigmoid; (g) Descending Colon; (h) Splenic Flexure; (i)-(l) Example of Colon Deformation.}
    \label{fig:endoscopic_image}
\end{figure}

{\color{black} Although the proximal-insertion endoscope does not achieve successful RL-based navigation, manual teleoperation can complete the corresponding navigation process in the same environment. The manual operation screen shots are shown in Fig.~\ref{fig:endoscopic_image} (c)-(d). This indicates that the simulated robot and environment are physically navigable, while current RL baselines fail to learn a robust proximal-insertion policy. The main difficulty is that proximal insertion does not directly command distal tip motion; distal advancement must emerge through shaft compliance, wall contact, friction, and colon deformation. In sharply curved and high-contact regions such as the sigmoid colon and splenic flexure, this creates a difficult long-horizon credit-assignment and recovery problem for RL.}

{\color{black}To directly assess deformable contact effects, we trained constrained PPO in a rigid colon and transferred the policy without adaptation to the matched deformable environment. On Task 2, the rigid-trained policy achieved 0 success, 2.36\% mean normalized progress, and a 100\% timeout rate, despite $s_c=0.707$, demonstrating a clear rigid-to-deformable transfer gap and further highlighting the necessity of a deformable benchmark for evaluating contact-rich colonoscopy navigation.}

\subsubsection{Reward Ablation Study}
{\color{black}We conduct the reward ablation using the magnetic-driven endoscope on Colon~1 and Colon~2 across Task~1--Task~4. These two anatomies are selected to evaluate the reward components under different geometric configurations and navigation difficulties. We do not include all anatomies in the ablation study because the remaining colon models are used for the cross-colon transfer evaluation in Section~IV-C, where they are used for testing policy transfer. Six reward variants are evaluated, including the center-alignment term $s_c$, the depth-existence term $s_1$, the depth-area term $s_2$, the lumen-visibility score $s_o$, the full reward $R$, and constrained PPO with the full reward $R$. Fig.~\ref{fig:test} reports the mean center-alignment score $s_c$ for each task on Colon~1 and Colon~2.}

{\color{black}Constrained PPO with the full reward $R$ achieves the best performance across all tasks, with the averaged $s_c$ over Colon~1 and Colon~2 being $s_c=0.882$, $0.898$, $0.919$, and $0.814$ for Task~1--4, respectively. In contrast, the individual reward components show inconsistent performance. The $s_c$-only reward performs moderately on Task~1--3, but degrades in Task~4. The $s_1$ and $s_o$ variants improve Task~3, but remain limited in the most difficult case. The $s_2$ variant gives the strongest single-component result in Task~4, but is weaker on Task~3.}

{\color{black} Constraint is important for stabilizing the full reward. Without constraint, the full reward performs well on Task~1 and Task~3, but drops on Task~2 and Task~4. Using constraints, all four tasks improve, especially Task~4. {\color{black}These results indicate that combining \(s_c\) with \(s_o\) provides the most consistent center-alignment results across the tested anatomies.}}

{\color{black}\subsubsection{Material Parameter Sensitivity Experiment}\label{sec:material_sensitivity}
{\color{black}To assess test-time robustness, we evaluated the fixed constrained PPO policy on Colon 1 under six material configurations, with \(E\in\{0.5,5,10\}\,\mathrm{MPa}\) and \(\nu\in\{0.45,0.49\}\). No retraining was performed, so this experiment evaluates policy robustness to material perturbations rather than their effect on RL training.}

Table~\ref{tab:material_sensitivity} reports the average center-alignment score and ROI alignment rate over Tasks~1--4. The mean $s_c$ remains stable across all settings, ranging from $0.826$ to $0.855$, indicating consistent lumen-centering behavior. The ROI rate varies moderately from $0.55$ to $0.70$, with slightly higher values for $\nu=0.45$ than for $\nu=0.49$. These results suggest that the overall visual navigation trend is not an artifact of a single Young's modulus or Poisson's ratio, although local ROI alignment is moderately affected by material properties.}

{\color{black}
\subsubsection{Anchor-Point Sensitivity Test}\label{sec:anchor_sensitivity}

We evaluated anchor-point sensitivity by testing the same constrained PPO checkpoint on Colon~1 under three fixation settings: sparse, default, and dense. The sparse setting used approximately 50\% of the default anchors, while the dense setting used approximately 175\% by adding neighboring simulation nodes around the default anchors. All other settings, including the camera, light, action scale, reward, material parameters, colon geometry, and task definitions, were unchanged.

Table~\ref{tab:anchor_sensitivity_pertask} reports the per-task results. Anchor sparsity mainly affects Task~2, the sigmoid-colon task, where the success rate drops from 100.0\% under the default setting to 16.7\%, and the reward decreases from 0.534 to $-0.377$. {\color{black}Overall, anchor-point sensitivity is task dependent, with no single fixation density consistently optimal across all tasks. The degradation of Task 2 under sparse anchoring indicates that sigmoid navigation is particularly sensitive to the fixation configuration, which remains a limitation of the current benchmark.}}

\begin{table}[t]
\centering
\begingroup
\small
\caption{Material sensitivity on Colon~1. Results are averaged over Tasks~1--4.}
\label{tab:material_sensitivity}
\begin{tabular*}{\columnwidth}{@{\extracolsep{\fill}}ccccc@{}}
\toprule
$E$ (MPa) & Poisson Ratio $\nu$ & Mean $s_c$ Score & ROI \\
\midrule
0.5  & 0.45 & 0.855 & 0.700 \\
0.5  & 0.49 & 0.846 & 0.600 \\
5.0  & 0.45 & 0.832 & 0.600 \\
5.0  & 0.49 & 0.826 & 0.550 \\
10.0 & 0.45 & 0.837 & 0.600 \\
10.0 & 0.49 & 0.830 & 0.550 \\
\bottomrule
\end{tabular*}
\endgroup
\end{table}

\begin{table}[t]
\centering
\begingroup
\color{black}
\small
\caption{Per-task anchor-point sensitivity results. Task~2 corresponds to sigmoid-colon navigation.}
\label{tab:anchor_sensitivity_pertask}
\begin{tabular*}{\columnwidth}{@{\extracolsep{\fill}}lccc|ccc@{}}
\toprule
\multirow{2}{*}{Task} 
& \multicolumn{3}{c|}{SR(\%)} 
& \multicolumn{3}{c}{Mean Step Reward} \\
\cmidrule(lr){2-4}\cmidrule(lr){5-7}
& Sparse & Default & Dense & Sparse & Default & Dense \\
\midrule
T1 & 100.0 & 100.0 & 100.0 & 0.593  & 0.696 & 0.695 \\
T2 & 16.7  & 30.0 & 80.0  & -0.377 & 0.534 & 0.195 \\
T3 & 100.0 & 90.0  & 60.0  & 0.672  & 0.593 & 0.677 \\
T4 & 100.0 & 80.0 & 100.0 & 0.589  & 0.622 & 0.590 \\
\bottomrule
\end{tabular*}
\endgroup
\end{table}

\subsection{Cross-Colon Policy Generalization}
\label{subsec:cross_colon}
{\color{black}
In this section, we evaluate cross-colon transfer by testing whether the trained policy can be deployed to different colon geometries. Since constrained PPO achieves the most stable performance under the default training setting, we use the constrained PPO policy trained on Colon~1 and Colon~2 as the representative transfer policy. To obtain a more reliable estimate than a single-run evaluation, the policy is tested on the rest three colons, with two repeated evaluations for each colon. Therefore, each colon-task pair is evaluated over ten trials in total.}

{\color{black}Transfer performance is summarized in Table~\ref{tab:cross-colon}. The trained constrained PPO exhibits limited cross-colon generalization. With distal-tip actuation, the policy achieves consistent success on T1 except in Colon 5 and shows moderate transfer on T3, whereas it largely fails on T2 and T4. With proximal-insertion actuation, the policy fails across all evaluated tasks, reinforcing it as the primary generalization bottleneck. Overall, T2 and T4 remain the most difficult transfer tasks, while T1 and T3 show partial transfer.}

\begin{table}[t!]
\begingroup
\small
\centering
\caption{{\color{black}Cross-colon test using constrained PPO with two flexible colonoscopes. ``T'' is short for ``Task''. \protect\ding{51} represents all trials succeed, and ``-'' represents all trials fail.}}
\label{tab:cross-colon}
\resizebox{\columnwidth}{!}{%
\begin{tabular}{c|cccc|c}
\toprule
Colon 
& \multicolumn{4}{c|}{Magnetic-driven} 
& Proximal-insertion\\
\cmidrule(lr){2-6}
& T1 & T2 & T3 & T4 & All Tasks \\
\midrule

Colon 3
&\ding{51} & 10\% & 50\% & 20\%
& - \\

Colon 4
&\ding{51} & 20\% & 70\% & 20\%
& - \\

Colon 5
& - & - & 50\% & -
& - \\

\bottomrule
\end{tabular}
}
\endgroup
\end{table}

\section{Conclusion}

This paper presents ARCGym, a reinforcement learning benchmark for image-based autonomous robotic colonoscopy in deformable anatomy. Built on Isaac Sim with GPU-accelerated soft-body physics, ARCGym provides clinically derived deformable colon models, three endoscope robots including a 6-DoF capsule, a magnetic-driven flexible endoscope, and a conventional flexible endoscope, and anatomically defined navigation tasks with a Gym-style interface.

{\color{black}
While the current benchmark uses reaching and passing the splenic flexure as a representative milestone for the main left-colon insertion bottlenecks, ARCGym is intended to enable reproducible, scalable evaluation of learning-based colonoscopy navigation and to facilitate progress on robust and safe long-horizon control under realistic deformation. Future work will extend the benchmark toward complete cecal intubation, additional anatomies, and sensing modalities, such as contact force measurement and shape sensing of the robot, to improve sim2real transfer and standardized safety metrics that better capture tissue interaction forces and clinical constraints.}

\section*{Acknowledgment}
This work is funded by the European Union, grant number 101135082. Views and opinions expressed are however those of the author(s) only and do not necessarily reflect those of the European Union or the European Health and Digital Executive Agency. Neither the European Union nor the granting authority can be held responsible for them. Hang Yin acknowledges the support from the NVIDIA Academic Grant Program. The authors thank Dr. Hans Stephensen for software assistance.

\bibliographystyle{unsrt}
\bibliography{references}

@article{brenner2024reduction,
  title={Reduction in colorectal cancer incidence by screening endoscopy},
  author={Brenner, Hermann and Heisser, Thomas and Cardoso, Rafael and Hoffmeister, Michael},
  journal={Nature Reviews Gastroenterology \& Hepatology},
  volume={21},
  number={2},
  pages={125--133},
  year={2024},
  publisher={Nature Publishing Group UK London}
}

@inproceedings{pore2022colonoscopy,
  title={Colonoscopy navigation using end-to-end deep visuomotor control: A user study},
  author={Pore, Ameya and Finocchiaro, Martina and Dall'Alba, Diego and Hernansanz, Albert and Ciuti, Gastone and Arezzo, Alberto and Menciassi, Arianna and Casals, Alicia and Fiorini, Paolo},
  booktitle={2022 IEEE/RSJ International Conference on Intelligent Robots and Systems (IROS)},
  pages={9582--9588},
  year={2022},
  organization={IEEE}
}

@inproceedings{corsi2023constrained,
  title={Constrained reinforcement learning and formal verification for safe colonoscopy navigation},
  author={Corsi, Davide and Marzari, Luca and Pore, Ameya and Farinelli, Alessandro and Casals, Alicia and Fiorini, Paolo and Dall'Alba, Diego},
  booktitle={2023 IEEE/RSJ International Conference on Intelligent Robots and Systems (IROS)},
  pages={10289--10294},
  year={2023},
  organization={IEEE}
}

@article{tan2025safe,
  title={Safe Navigation for Robotic Digestive Endoscopy via Human Intervention-based Reinforcement Learning},
  author={Tan, Min and Tao, Yushun and Zheng, Boyun and Xie, Gaosheng and Feng, Lijuan and Xia, Zeyang and Xiong, Jing},
  journal={Expert Systems with Applications},
  pages={128841},
  year={2025},
  publisher={Elsevier}
}

@inproceedings{xu2021surrol,
  title={Surrol: An open-source reinforcement learning centered and dvrk compatible platform for surgical robot learning},
  author={Xu, Jiaqi and Li, Bin and Lu, Bo and Liu, Yun-Hui and Dou, Qi and Heng, Pheng-Ann},
  booktitle={2021 IEEE/RSJ International Conference on Intelligent Robots and Systems (IROS)},
  pages={1821--1828},
  year={2021},
  organization={IEEE}
}

@inproceedings{schmidgall2024surgical,
  title={Surgical Gym: A high-performance GPU-based platform for reinforcement learning with surgical robots},
  author={Schmidgall, Samuel and Krieger, Axel and Eshraghian, Jason},
  booktitle={2024 IEEE International Conference on Robotics and Automation (ICRA)},
  pages={13354--13361},
  year={2024},
  organization={IEEE}
}

@article{wu2024surgicai,
  title={Surgicai: A hierarchical platform for fine-grained surgical policy learning and benchmarking},
  author={Wu, Jin and Zhou, Haoying and Kazanzides, Peter and Munawar, Adnan and Liu, Anqi},
  journal={Advances in Neural Information Processing Systems},
  volume={37},
  pages={63771--63789},
  year={2024}
}

@article{jianu2024cathsim,
  title={CathSim: an open-source simulator for endovascular intervention},
  author={Jianu, Tudor and Huang, Baoru and Vu, Minh Nhat and Abdelaziz, Mohamed EMK and Fichera, Sebastiano and Lee, Chun-Yi and Berthet-Rayne, Pierre and y Baena, Ferdinando Rodriguez and Nguyen, Anh},
  journal={IEEE Transactions on Medical Robotics and Bionics},
  volume={6},
  number={3},
  pages={971--979},
  year={2024},
  publisher={IEEE}
}

@INPROCEEDINGS{11127627,
  author={Yao, Tianliang and Ban, Madaoji and Lu, Bo and Pei, Zhiqiang and Qi, Peng},
  booktitle={2025 IEEE International Conference on Robotics and Automation (ICRA)}, 
  title={Sim4EndoR: A Reinforcement Learning Centered Simulation Platform for Task Automation of Endovascular Robotics}, 
  year={2025},
  volume={},
  number={},
  pages={824-830},
  doi={10.1109/ICRA55743.2025.11127627}}

@inproceedings{varier2022ambf_rl,
  title={AMBF-RL: A real-time simulation based reinforcement learning toolkit for medical robotics},
  author={Varier, Vignesh Manoj and Rajamani, Dhruv Kool and Tavakkolmoghaddam, Farid and Munawar, Adnan and Fischer, Gregory S},
  booktitle={2022 International Symposium on Medical Robotics (ISMR)},
  pages={1--8},
  year={2022},
  organization={IEEE},
  doi={10.1109/ISMR48347.2022.9807609}
}

@article{zhao2024bronchocopilot,
  title={BronchoCopilot: Towards autonomous robotic bronchoscopy via multimodal reinforcement learning},
  author={Zhao, Jianbo and Chen, Hao and Tian, Qingyao and Chen, Jian and Yang, Bingyu and Liu, Hongbin},
  journal={arXiv preprint arXiv:2403.01483},
  year={2024}
}

@article{yu2024orbit,
  title={ORBIT-Surgical: An Open-Simulation Framework for Learning Surgical Augmented Dexterity},
  author={Yu, Qinxi and Moghani, Masoud and Dharmarajan, Karthik and Schorp, Vincent and Panitch, William Chung-Ho and Liu, Jingzhou and Hari, Kush and Huang, Huang and Mittal, Mayank and Goldberg, Ken and Garg, Animesh},
  journal={arXiv preprint arXiv:2404.16027},
  year={2024}
}

@article{richter1903open,
  title={Open-sourced reinforcement learning environments for surgical robotics (2019)},
  author={Richter, Florian and Orosco, Ryan K and Yip, M},
  journal={arXiv preprint arXiv:1903.02090}
}

@InProceedings{endovla2025corl,
  title     = {EndoVLA: Dual-Phase Vision-Language-Action for Precise Autonomous Tracking in Endoscopy},
  author    = {KIT, NG CHI and Bai, Long and Wang, Guankun and Wang, Yupeng and Gao, Huxin and yuan, Kun and Jin, Chenhan and Zeng, Tieyong and Ren, Hongliang},
  booktitle = {Proceedings of The 9th Conference on Robot Learning},
  pages     = {4958--4974},
  year      = {2025},
  editor    = {Lim, Joseph and Song, Shuran and Park, Hae-Won},
  volume    = {305},
  series    = {Proceedings of Machine Learning Research},
  month     = {27--30 Sep},
  publisher = {PMLR},
  url       = {https://proceedings.mlr.press/v305/kit25a.html}
}

@article{turan2019learning,
  title   = {Learning to navigate endoscopic capsule robots},
  author  = {Turan, Mehmet and Almalioglu, Yasin and Gilbert, Hunter B and Mahmood, Faisal and Durr, Nicholas J and Araujo, Helder and Sar{\i}, Alp Eren and Ajay, Anurag and Sitti, Metin},
  journal = {IEEE Robotics and Automation Letters},
  volume  = {4},
  number  = {3},
  pages   = {3075--3082},
  year    = {2019}
}

@article{li2023rl,
  title={Rl-tee: Autonomous probe guidance for transesophageal echocardiography based on attention-augmented deep reinforcement learning},
  author={Li, Keyu and Li, Ang and Xu, Yangxin and Xiong, Huahua and Meng, Max Q-H},
  journal={IEEE Transactions on Automation Science and Engineering},
  volume={21},
  number={2},
  pages={1526--1538},
  year={2023},
  publisher={IEEE}
}

@article{tan2019laparoscopy_drl,
  author  = {Tan, Xiaoyu and Chng, Chin{-}Boon and Su, Ye and Lim, Kah{-}Bin and Chui, Chee{-}Kong},
  title   = {Robot-Assisted Training in Laparoscopy Using Deep Reinforcement Learning},
  journal = {{IEEE} Robotics Autom. Lett.},
  volume  = {4},
  number  = {2},
  pages   = {485--492},
  year    = {2019},
  doi     = {10.1109/LRA.2019.2891311},
  url     = {https://doi.org/10.1109/LRA.2019.2891311}
}

@article{ji2021towards,
  title={Towards safe control of continuum manipulator using shielded multiagent reinforcement learning},
  author={Ji, Guanglin and Yan, Junyan and Du, Jingxin and Yan, Wanquan and Chen, Jibiao and Lu, Yongkang and Rojas, Juan and Cheng, Shing Shin},
  journal={IEEE Robotics and Automation Letters},
  volume={6},
  number={4},
  pages={7461--7468},
  year={2021},
  publisher={IEEE}
}

@article{scheikl2023lapgym,
  title={Lapgym-an open source framework for reinforcement learning in robot-assisted laparoscopic surgery},
  author={Scheikl, Paul Maria and Gyenes, Bal{\u{A}}{\k{A}}zs and Younis, Rayan and Haas, Christoph and Neumann, Gerhard and Wagner, Martin and Mathis-Ullrich, Franziska},
  journal={Journal of Machine Learning Research},
  volume={24},
  number={368},
  pages={1--42},
  year={2023}
}

@inproceedings{zhang2022deep,
  title={Deep reinforcement learning-based control for stomach coverage scanning of wireless capsule endoscopy},
  author={Zhang, Yameng and Bai, Long and Liu, Li and Ren, Hongliang and Meng, Max Q--H},
  booktitle={2022 IEEE International Conference on Robotics and Biomimetics (ROBIO)},
  pages={01--06},
  year={2022},
  organization={IEEE}
}

@article{long2025surgical,
  title={Surgical embodied intelligence for generalized task autonomy in laparoscopic robot-assisted surgery},
  author={Long, Yonghao and Lin, Anran and Kwok, Derek Hang Chun and Zhang, Lin and Yang, Zhenya and Shi, Kejian and Song, Lei and Fu, Jiawei and Lin, Hongbin and Wei, Wang and others},
  journal={Science Robotics},
  volume={10},
  number={104},
  pages={eadt3093},
  year={2025},
  publisher={American Association for the Advancement of Science}
}

@article{martin2020enabling,
  title={Enabling the future of colonoscopy with intelligent and autonomous magnetic manipulation},
  author={Martin, James W and Scaglioni, Bruno and Norton, Joseph C and Subramanian, Venkataraman and Arezzo, Alberto and Obstein, Keith L and Valdastri, Pietro},
  journal={Nature machine intelligence},
  volume={2},
  number={10},
  pages={595--606},
  year={2020},
  publisher={Nature Publishing Group UK London}
}

@article{massalou2019mechanical,
  title={Mechanical effects of load speed on the human colon},
  author={Massalou, D and Masson, C and Afquir, S and Baqu{\'e}, P and Arnoux, P-J and B{\`e}ge, T},
  journal={Journal of Biomechanics},
  volume={91},
  pages={102--108},
  year={2019},
  publisher={Elsevier}
}

@article{cheng2013modeling,
  title={Modeling and in vitro experimental validation for kinetics of the colonoscope in colonoscopy},
  author={Cheng, Wu-Bin and Di, Yun-Yun and Zhang, Edwin M and Moser, Michael AJ and Kanagaratnam, Sivaruban and Korman, Louis Y and Sarvazyan, Noune and Zhang, Wen-Jun},
  journal={Annals of biomedical engineering},
  volume={41},
  number={5},
  pages={1084--1093},
  year={2013},
  publisher={Springer}
}

@book{cohen2022successful,
  title={Successful training in gastrointestinal endoscopy},
  author={Cohen, Jonathan},
  year={2022},
  publisher={John Wiley \& Sons}
}

@article{finocchiaro2026clinically,
  title={Clinically validated dataset of 435 human colons segmented from CT colonography},
  author={Finocchiaro, Martina and Stern, Ronja and Vilhelmsborg, Rikke and Smith, Abraham George and Petersen, Jens and Cold, Kristoffer and Konge, Lars and Erleben, Kenny and Ganz, Melanie},
  journal={Scientific Data},
  year={2026},
  publisher={Nature Publishing Group UK London}
}

@article{loeve2013mechanical,
  title={Mechanical Analysis of Insertion Problems and Pain During Colonoscopy: Why Highly Skill-Dependent Colonoscopy Routines are Necessary in the First Place… and How They May be Avoided},
  author={Loeve, Arjo J and Fockens, Paul and Breedveld, Paul},
  journal={Canadian Journal of Gastroenterology and Hepatology},
  volume={27},
  number={5},
  pages={293--302},
  year={2013},
  publisher={Wiley Online Library}
}

@article{Tian2023BowelCancer,
  author    = {J. Tian and K. O. Afebu and A. Bickerdike and others},
  title     = {Fundamentals of Bowel Cancer for Biomedical Engineers},
  journal   = {Annals of Biomedical Engineering},
  volume    = {51},
  pages     = {679--701},
  year      = {2023},
  month     = apr,
  doi       = {10.1007/s10439-023-03155-8},
  publisher = {Springer}
}

@ARTICLE{6041014,
  author={Ciuti, Gastone and Menciassi, Arianna and Dario, Paolo},
  journal={IEEE Reviews in Biomedical Engineering}, 
  title={Capsule Endoscopy: From Current Achievements to Open Challenges}, 
  year={2011},
  volume={4},
  number={},
  pages={59-72},
  doi={10.1109/RBME.2011.2171182}}

@article{jones2006kinematics,
  title={Kinematics for multisection continuum robots},
  author={Jones, Bryan A and Walker, Ian D},
  journal={IEEE Transactions on Robotics},
  volume={22},
  number={1},
  pages={43--55},
  year={2006},
  publisher={IEEE}
}

@article{waye2013difficult,
  title={Difficult colonoscopy},
  author={Waye, Jerome D},
  journal={Gastroenterology \& hepatology},
  volume={9},
  number={10},
  pages={676},
  year={2013}
}

@article{mccarthy2016early,
  title={Early splenic flexure intubation competency predicts early cecal intubation competency in gastroenterology fellows},
  author={McCarthy, Sean T and Jorgensen, Jennifer and Elta, Grace H and Kolars, Joseph C and Korsnes, Sheryl and Metko, Valbona and Stout, James and Rubenstein, Joel H},
  journal={Digestive diseases and sciences},
  volume={61},
  number={11},
  pages={3155--3160},
  year={2016},
  publisher={Springer}
}

@inproceedings{yang2024depth,
  title={Depth anything: Unleashing the power of large-scale unlabeled data},
  author={Yang, Lihe and Kang, Bingyi and Huang, Zilong and Xu, Xiaogang and Feng, Jiashi and Zhao, Hengshuang},
  booktitle={Proceedings of the IEEE/CVF conference on computer vision and pattern recognition},
  pages={10371--10381},
  year={2024}
}

\end{document}